\documentclass[10pt,twocolumn,letterpaper]{article}

\usepackage{cvpr}
\usepackage{times}
\usepackage{epsfig}
\usepackage{graphicx}
\usepackage{amsmath}
\usepackage{amssymb}
\usepackage{verbatim}
\usepackage{wrapfig}
\usepackage{animate}
\usepackage{multirow}

\usepackage{color}
\usepackage{subfig}
\usepackage{xspace}
\cvprfinalcopy

\usepackage[pagebackref=true,breaklinks=true,letterpaper=true,colorlinks,bookmarks=false]{hyperref}

\def\cvprPaperID{XXXX} 

\ifcvprfinal\fi
\begin{document}

\title{Deep Learning with Energy-efficient Binary Gradient Cameras}

\author{Suren Jayasuriya$^{*,\ddagger}$,~~~Orazio Gallo$^*$,~~~Jinwei Gu$^*$,~~~Jan Kautz$^*$\\
$^*$NVIDIA,~~~~~$^\ddagger$Carnegie Mellon University\\
}

\maketitle
\thispagestyle{empty}

\begin{abstract}
Power consumption is a critical factor for the deployment of embedded computer vision systems.
We explore the use of computational cameras that directly output binary gradient images to reduce the portion of the power consumption allocated to image sensing. We survey the accuracy of binary gradient cameras on a number of computer vision tasks using deep learning. These include object recognition, head pose regression, face detection, and gesture recognition. We show that, for certain applications, accuracy can be on par or even better than what can be achieved on traditional images. 
We are also the first to recover intensity information from binary spatial gradient images---useful for applications with a human observer in the loop, such as surveillance.
Our results, which we validate with a prototype binary gradient camera, point to the potential of gradient-based computer vision systems.
\end{abstract}

\vspace{-5mm}
\section{Introduction}\label{sec:intro}

Recent advances in deep learning have significantly improved the accuracy of computer vision tasks such as visual recognition, object detection, segmentation, and others.  Leveraging large datasets of RGB images and GPU computation, many of these algorithms now match, or even surpass, human performance. This accuracy increase makes it possible to deploy these computer vision algorithms in the wild. Power consumption, however, remains a critical factor for embedded and mobile applications, where battery life is a key design constraint.

\begin{figure}
\centering
\includegraphics[width=\columnwidth]{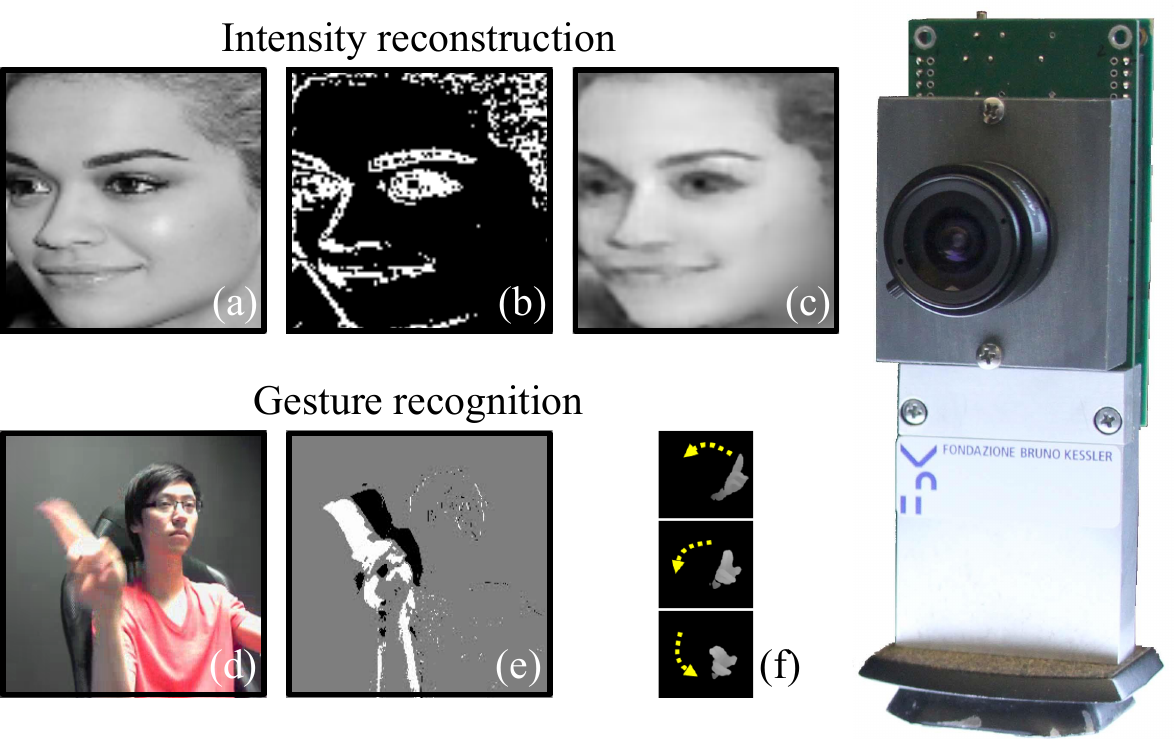}
\caption{Two of the tasks we study in the context of binary gradient images. Insets (a) and (d) are traditional pictures of the scene. Inset (b) is a simulated, spatial binary gradient, and (e) a simulated temporal binary gradient. From these we can reconstruct the original intensity image (c) or perform gesture recognition (f). We also used real data captured with the prototype shown on the right. Inset (f) is from Molchanov \etal~\cite{molchanov2016nvgesture}.}\label{fig:teaser}
\end{figure}

For instance, Google Glass operating a modern face recognition algorithm has a battery life of less than 40 minutes, with image sensing and computation each consuming roughly 50\% of the power budget~\cite{likamwa2014}. Moreover, research in computer architecture has focused on energy-efficient accelerators for deep learning, which reduce the power footprint of neural network inference to the mW range~\cite{eie, truenorth}, bringing them in the same range of the power consumption as image sensing.

When the computer vision algorithms are too computationally intensive, or would require too much power for the embedded system to provide, the images can be uploaded to the cloud for off-line processing. However, even when using image or video compression, the communication cost can still be prohibitive for embedded systems, sometimes by several orders of magnitude~\cite{ragan2014phd}. Thus an image sensing strategy that reduces the amount of captured data can have an impact on the overall power consumption that extends beyond just acquisition and processing.

\begin{figure}
\centering
\includegraphics[width=.49\columnwidth, height=.8in]{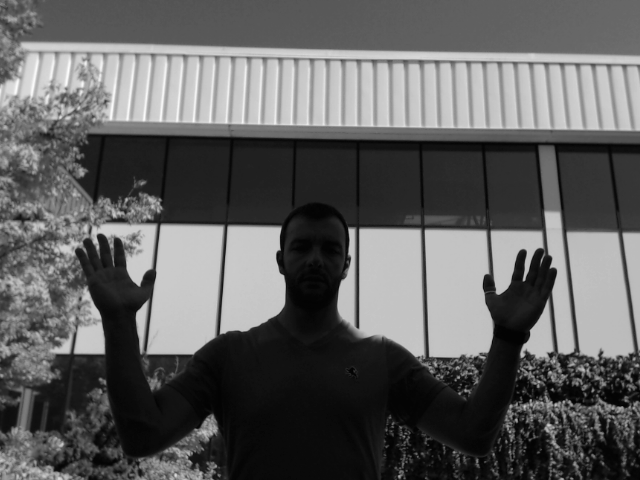}
\includegraphics[width=.49\columnwidth]{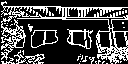}
\caption{A traditional image (left) and an example of real spatial binary gradient data (right). Note that these pictures were taken with different cameras and lenses and, thus, do not exactly match.}\label{fig:spatialgradcameras}
\end{figure}

A large component of the image sensing power is burned to capture dense images or videos, meaning that each pixel is associated with a value of luminance, color component, depth, or other physical measurement. Not all pixels, however, carry valuable information: pixels capturing edges tend to be more informative than pixels in flat areas.  Recently, novel sensors have been used to feed gradient data directly to the computer vision algorithms.~\cite{chen2016, weikersdorfer2014}. In addition, there has been a growing interested in event based cameras such as those proposed by Lichsteiner \etal~\cite{dvs}. These cameras consume significantly less power than traditional cameras, and record binary changes of illumination at the pixel level, and only output pixels when they become active. Another particularly interesting type of sensor was proposed by Gottardi \etal~\cite{graincam}. This sensor produces a binary image where only the pixels in high-gradient regions become active; depending on the modality of operation, only active pixels, or pixels that changed their activity status between consecutive frames, can then be read. The resulting images appear like binary edge images, see Figure~\ref{fig:spatialgradcameras}.

While these designs allow for a significant reduction of the power required to acquire, process, and transmit images, it also limits the information that can be extracted from the scene. The question, then, becomes whether this results in a loss of accuracy for the computer vision algorithms, and if such loss is justified by the power saving.

\subsection{Our Contributions}

In this paper, we focus on two aspects related to the use of binary gradient cameras for low-power, embedded computer vision applications.

First, we explore the tradeoff between energy and accuracy this type of data introduces on a number of computer vision tasks. To avoid having to hand-tune traditional computer vision algorithms to binary gradient data, we use deep learning approaches as benchmarks, and leverage the networks' ability to learn by example. We select a number of representative tasks, and analyze the change in accuracy of established neural network-based approaches, when they are applied to binarized gradients. 

Second, we investigate whether the intensity information can be reconstructed from these images in post-processing, for those tasks where it would be useful for a human to visually inspect the captured image, such as long-term video surveillance on a limited power budget. Unlike other types of gradient-based sensors, intensity reconstruction is an ill-posed problem for our type of data because both the direction and the sign of the gradient are lost, see Section~\ref{sec:reconstruction}. To the best of our knowledge, in fact, we are the first to show intensity reconstruction from single-shot, spatial binary gradients.

We perform our formal tests simulating the output of the sensor on existing datasets, but we also validate our findings by capturing real data with the prototype developed by Gottardi \etal~\cite{graincam} and described in Section~\ref{sec:operation}. 

We believe that this paper presents a compelling reason for using binary gradient cameras in certain computer vision tasks, to reduce the power consumption of embedded systems.

\section{Related Work}\label{sec:related}

We describe the prior art in terms of the gradient cameras that have been proposed, and then in terms of computer vision algorithms developed for this type of data.

\textbf{Gradient cameras} can compute spatial gradients either in the optical domain~\cite{chen2016,zomet2006lensless,koppal2013toward}, or on-board the image sensor, a technique known as focal plane processing~\cite{chai2000focal,leon2007focalCmos,nilchi2009focal,hasler2005low}. The gradients can be either calculated using adjacent pixels~\cite{graincam} or using current-mode image sensors~\cite{gruev2004currentmode}. Some cameras can also compute temporal gradient images, \ie images where the active pixels indicate a temporal change in local contrast~\cite{graincam,dvs}. Most of these gradient cameras have side benefits of fast frame rates and reduced data bandwidth/power due to the sparseness of gradients in a scene. In fact, the camera by Lichtsteiner \etal can read individual pixels when they become active~\cite{dvs}. Moreover, the fact that gradient cameras output a function of the difference of two or more pixels, rather than the pixel values themselves, allows them to deal with high-dynamic-range scenes.

\textbf{Applications of gradient cameras} were first exposited in the work by Tumblin \etal, who described the advantages of reading pixel differences rather than absolute values~\cite{tumblin2005gradient}. A particular area of interest for temporal binary gradients and event-based cameras is SLAM (simultaneous localization and mapping) and intensity reconstruction. Researchers have shown SLAM~\cite{weikersdorfer2013simultaneous}, simultaneous intensity reconstruction and object tracking~\cite{kim2008simultaneous}, combined optical flow and intensity reconstruction~\cite{bardow2016}, and simultaneous depth, localization, and intensity reconstruction~\cite{kim2016real}. In addition, some early work has focused on using spiking neural networks for event-based cameras~\cite{oconnor2015spiking}. The common denominator to all of these techniques is that the camera, or at least the scene, must be dynamic: the sensor does not output any information otherwise. For tradeoffs between energy and visual recognition accuracy, recent work proposed optically computing the first layer of convolutional neural networks using Angle Sensitive Pixels~\cite{chen2016}. However, the camera required slightly out-of-focus scenes to perform this optical convolution and did not work with binary gradient images.

In our work, we focus on the camera proposed by Gottardi \etal~\cite{graincam}, which can produce spatial binary gradients, and can image static scenes as well as dynamic ones.  Gasparini \etal showed that this camera can be used as a long-lifetime node in wireless networks~\cite{gasparini2011ultralow}. This camera was also used to implement low-power people counter~\cite{gasparini2010counter}, but only in the temporal gradient modality (see Section~\ref{sec:operation}).

\section{Binary Gradient Cameras}

In this section, we define the types of binary gradient images we are considering and we analyze the power and high dynamic range benefits from such cameras.

\subsection{Operation}\label{sec:operation}

For spatial binary gradients, we refer to cameras where a pixel becomes active when a local measure of contrast is above threshold.
Specifically, for two pixels $i$ and $j$, we define the difference $\Delta_{i,j} = |I_i - I_j|$, where $I$  
\begin{wrapfigure}{r}{2cm}
\vspace{-7mm}
\begin{center}
\includegraphics[width=2cm]{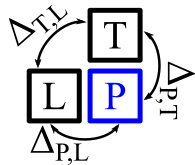}
\end{center}
\vspace{-2mm}
\vspace{-3.5mm}
\end{wrapfigure}
is the measured pixel's brightness. We also define a neighborhood $\nu$ consisting of pixel P and the pixels to its left, L, and top, T (see inset). The output at pixel P will then be:
\begin{equation}\label{eq:gs}
G_S(\text{P}) = 
\begin{cases}
1 & \text{if}~~\underset{i,j \in \nu}{\max}~~\Delta_{i,j} > T\\
0& \text{otherwise}
\end{cases},
\end{equation}
where $T$ is a threshold set at capture time. The output of this operation is a binary image where changes in local spatial contrast above threshold yield a 1, else a 0, see Figure~\ref{fig:spatialgradcameras}. Note that this operation is an approximation of a binary local derivative: $\Delta_{T,L}$ alone can trigger an activation for P, even though the intensity at P is not significantly different from either of the neighbors'. It can be shown that the consequence of this approximation is a ``fattening'' of the image edges by a factor of roughly $\sqrt{2}$ when compared to the magnitude of the a gradient computed with regular finite differences. The advantage of this formulation is that it can be implemented efficiently in hardware. 

For temporal binary gradients, the sensor proposed by Lichtsteiner \etal~\cite{dvs}, which works asynchronously, outputs +1 (-1) for a pixel whose intensity increases (decreases) by a certain threshold, and 0 otherwise. The sensor proposed by Gottardi \etal produces a slightly different image for temporal gradients, where the value of a pixel is the difference between its current and previous binary spatial gradient~\cite{graincam}:
\begin{equation}\label{eq:gt}
G_T(\text{P},t) = \max \left(0, | G_S(\text{P},t) - G_S(\text{P},t-1)|\right),
\end{equation}
where we made the dependency on time $t$ explicit. This is implemented by storing the previous value in a 1-bit memory collocated with the pixel to avoid unnecessary data transfer. An image produced by this modality can be seen in Figure~\ref{fig:teaser}(e).

\subsection{Power Considerations}
Binary gradient cameras have numerous advantages in terms of power and bandwidth.
A major source of power consumption in modern camera sensors is the analog-to-digital conversion and the transfer of the 12-16 bits data off-chip, to subsequent image processing stages. 
Gradients that employ 1 or 2 bits can significantly reduce both the cost for the conversion, and the amount data to be encoded at the periphery of the array.  In fact, the sensor only transfers the addresses of the pixels that are active, and when no pixels are active, no power is used for transferring data.

Comparing power consumption for sensors of different size, technology, and mode of operation is not easy. Our task is further complicated by the fact that the power consumption for a binary gradient sensor is a function of the contrast in the scene. However, here we make some assumptions to get a very rough figure.  Gottardi \etal~\cite{graincam} report that the number of active pixels is usually below 25\% (in the data we captured, we actually measured that slightly less than 10\% of the pixels were active on average). The power consumption for the sensor by Gottardi \etal can be approximated by the sum of two components. The first, independent of the actual number of active pixels, is the power required to scan the sensor and amounts to 0.0024$\mu$W/pixel. The second is the power required to deliver the addresses of the active pixels, and is 0.0195$\mu$W/pixel~\cite{gallo2015gray}. At 30fps, this power corresponds to 7.3pJ/pixels. A modern image sensor, for comparison, is over 300pJ/pixel~\cite{asp}. Once again, these numbers are to be taken as rough estimates.

\section{Experiments}

In this section, we describe the vision tasks we used to benchmark spatial and temporal binary gradients. For the benchmarks involving static scenes or single images, we could only test spatial gradients. We used TensorFlow and Keras to construct our networks. All experiments were performed on a cluster of GPUs with NVIDIA Titan X's or K80s. For all the experiments in this section, we picked a reference baseline network appropriate for the task, we trained it on intensity or RGB images, and compared the performance of the same architecture on data that simulates the sensor by Gottardi \etal~\cite{graincam}. An example of such data can be seen in Figure~\ref{fig:teaser}(b) and \ref{fig:teaser}(c). Table~\ref{table:simresults} summarizes all the comparisons we describe below.

\subsection{Computer Vision Benchmarks}\label{sec:benchmarks}


\begin{table*}
\begin{center}
\begin{tabular}{ccccc}
\hline
Task & Dataset & Traditional & Binary gradient & Network used\\
\hline
\multirow{3}{*}{Recognition} & MNIST~\cite{mnist} & 99.19\% & 98.43\% & LeNet~\cite{lenet}\\ \cline{2-5}
& CIFAR-10~\cite{cifar10} & 77.01\% & 65.68\% & LeNet~\cite{lenet}\\ \cline{2-5}
& \multirow{2}{*}{NVGesture~\cite{molchanov2016nvgesture}} & \multirow{2}{*}{72.5\%} & $G_T$:~ 74.79\% & \multirow{2}{*}{Molchanov \etal~\cite{molchanov2016nvgesture}}\\ 
&&&$G_S$:~ 65.42\% & \\
\hline
\multirow{2}{*}{Head pose} & 300VW~\cite{300vw} & 0.6$^\circ$ & 1.8$^\circ$ & LeNet~\cite{lenet} \\ \cline{2-5}
& BIWI Face Dataset~\cite{biwi} & 3.5$^\circ$ & 4.3$^\circ$ & VGG16~\cite{vgg16}\\
\hline
\multirow{3}{*}{Face detection --- WIDER~\cite{wider}} &  Easy & 89.2\% &  74.5\% & \multirow{3}{*}{Faster R-CNN~\cite{renNIPS15fasterrcnn}} \\ \cline{2-4}
& Medium & 79.2\% & 60.5\% &  \\ \cline{2-4}
& Hard & 40.2\% & 28.3\% & \\ 
\hline
\end{tabular}
\end{center}
\vspace{-3mm}
\caption{Summary of the comparison between traditional images and binary gradient images on visual recognition tasks.}\label{table:simresults}
\end{table*}

\vspace{2mm}
\noindent\textbf{Object Recognition ---} We used MNIST~\cite{mnist} and CIFAR-10~\cite{cifar10} to act as common baselines, and for easy comparison with other deep learning architectures, on object recognition tasks. MNIST comprises 60,000, 28x28 images of handwritten digits. CIFAR-10 has 60,000, 32x32 images of objects from 10 classes, with 10,000 additional images for validation. For these tasks we used LeNet~\cite{lenet}.

On MNIST, using simulated binary gradient data degrades the accuracy by a mere 0.76\%. For CIFAR-10, we trained the baseline on RGB images. The same network, trained on the simulated data, achieves a loss in accuracy of 11.33\%. For reference, using grayscale instead of RGB images causes a loss of accuracy of 4.86\%, which is roughly comparable to the difference in accuracy between using grayscale and gradient images---but without the corresponding power saving.

\noindent\textbf{Head Pose Regression ---} We also explored single-shot head pose regression, an important use-case for human-computer interaction, and driver monitoring in vehicles. We used two datasets to benchmark the performance of gradient cameras on head pose regression. The first, the BIWI face dataset, contains 15,000 images of 20 subjects, each accompanied by a depth image, as well as the head 3D location and orientation~\cite{biwi}. The second, the 300VW dataset, is a collection of 300 videos of faces annotated with 68 landmark points~\cite{300vw}. We used the landmark points to estimate the head orientation.

On the BIWI dataset, training a LeNet from scratch did not yield network convergence. Therefore, we used a pretrained VGG16~\cite{vgg16} network on the RGB images. We then fine-tuned the network on the simulated binary gradient data. The network trained on simulated binary gradient data yields a degradation of estimation accuracy of a 0.8 mean degree error per pixel. On the  300VW dataset, we trained LeNet on the simulated data. The mean angular error increases by 1.2 degrees per pixel, which is small when accounting for the corresponding power saving.

\begin{figure}
\centering
\includegraphics[height=0.75in,trim={30px 0 0 0}, clip]{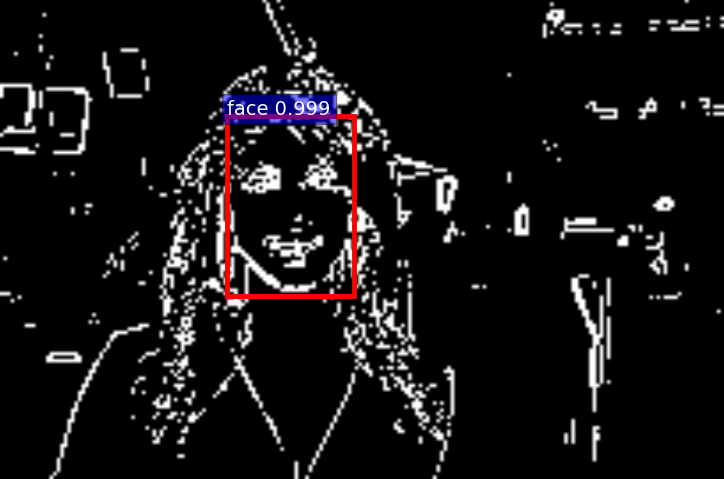}
\includegraphics[height=0.75in,trim={100px 0 0 0}, clip]{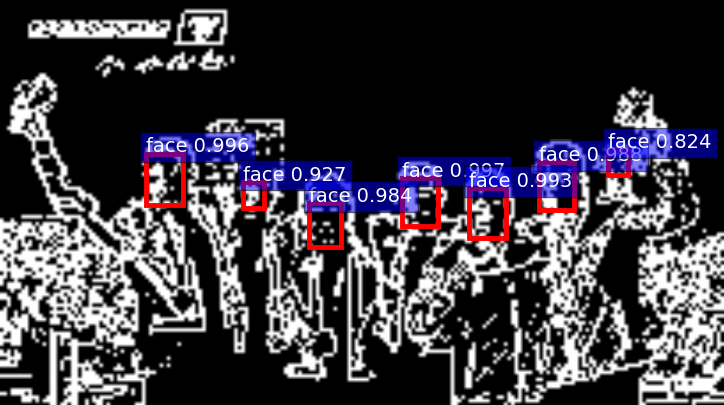}
\includegraphics[height=0.75in]{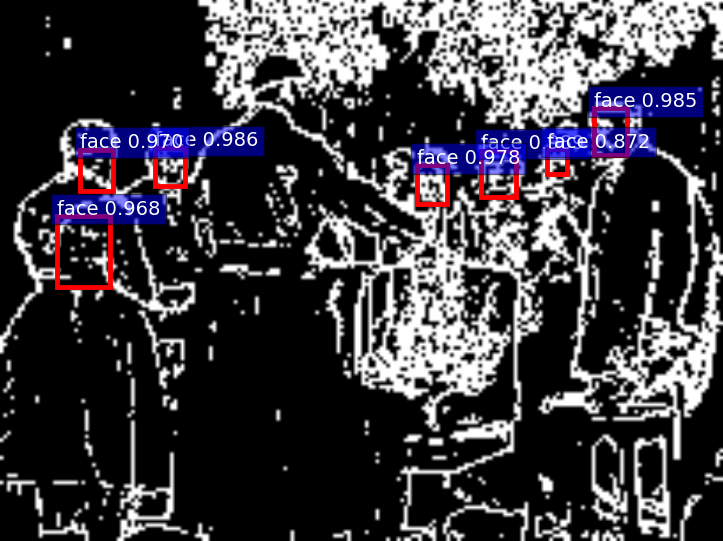}
\caption{Face detection on binary spatial gradient images simulated from the WIDER dataset.}\label{fig:facedetect}
\end{figure}

\noindent\textbf{Face Detection ---} Another traditional vision task is face detection. For this task we trained the network on the WIDER face dataset, a collection of 30,000+ images with 390,000+ faces, and is organized in three categories for face detection: easy, medium, and hard~\cite{wider}. Figure~\ref{fig:facedetect} shows representative images of different levels of difficulty. Note that this dataset is designed to be very challenging, and includes pictures taken under extreme scale, illumination, pose, and expression changes, among other factors.

For this task, we used the network proposed by Ren \etal~\cite{renNIPS15fasterrcnn}. Once again, we trained it on both the RGB and the simulated binary gradient images. The results are summarized in Table~\ref{table:simresults}. On this task, the loss in accuracy due to using the binary gradient data is more significant, ranging from 11.9\% to 18.7\%, depending on the category.

\noindent\textbf{Gesture Recognition ---} Our final task was gesture recognition. Unlike the previous benchmarks, whose task can be defined on a single image, this task has an intrinsic temporal component: the same hand position can be found in a frame extracted from two different gestures. Therefore, for this task we test both the spatial and temporal modalities.

\begin{figure}
\centering
\includegraphics[width=.31\columnwidth]{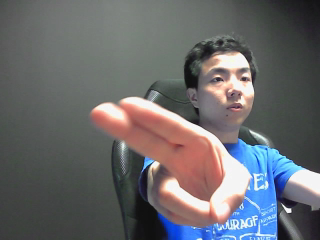}\enskip
\includegraphics[width=.31\columnwidth]{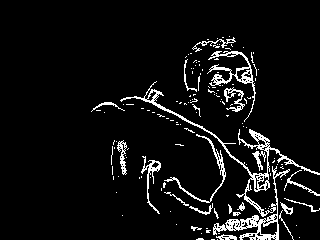}\enskip
\includegraphics[width=.31\columnwidth]{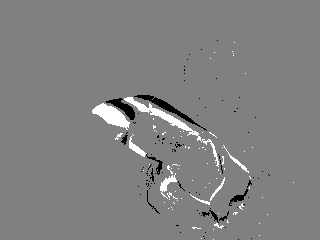}\\
\vspace{-3mm}
\subfloat[]{\includegraphics[width=.31\columnwidth]{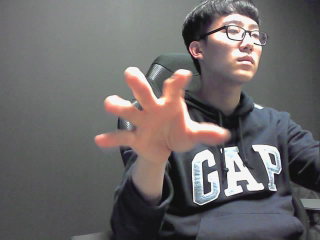}}\enskip
\subfloat[]{\includegraphics[width=.31\columnwidth]{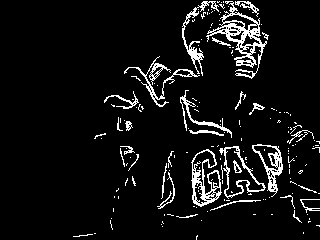}}\enskip
\subfloat[]{\includegraphics[width=.31\columnwidth]{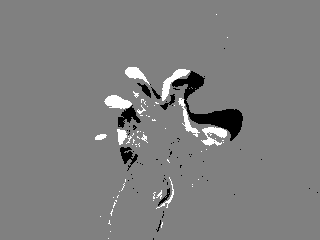}}\enskip
\caption{Two frames from the NVIDIA Dynamic Hand Gesture Dataset~\cite{molchanov2016nvgesture}, (a), the corresponding spatial binary gradients, (b), and temporal binary gradients, (c).}\label{fig:gesture}
\end{figure}

We used the dataset released by Molchanov \etal, which contains 1,500+ hand gestures from 25 gesture classes, performed by 20 different subjects~\cite{molchanov2016nvgesture}. The dataset offers several acquisition modalities, including RGB, IR, and depth, and was randomly split between training (70\%) and testing (30\%) by the authors. The network for this algorithm was based on~\cite{molchanov2016nvgesture}, which used an RNN on top of 3D convolutional features.  We limited our tests to RGB inputs, and did not consider the other types of data the dataset offers, see Figure~\ref{fig:gesture}. As shown in Table~\ref{table:simresults}, the simulated spatial binary gradient modality results in an accuracy degradation of 7.08\% relative to RGB images and 5.41\% relative to grayscale. However, as mentioned before, this task has a strong temporal component and one would expect that the temporal gradient input should perform better. Indeed, the temporal modality yields increased accuracy on both grayscale (+3.96\%) and RGB (+2.29\%) data. This is a significant result, because the additional accuracy is possible thanks to data that is actually cheaper to acquire from a power consumption standpoint. Note that the input to the network is a set of non-overlapping clips of 8 frames each, so the network can still ``see'' temporal information in modalities other than the temporal binary gradients.

Across a variety of tasks, we see that the accuracy on binary gradient information varies. It is sometimes comparable to, and sometimes better than, the accuracy obtained on traditional intensity data. Other times there is a significant accuracy loss is significant, as is the case with face detection. We think that this is due in part to the task, which can benefit from information that is lost in the binary gradient data, and in part to the challenging nature of the dataset.
Our investigation suggests that the choice of whether a binary gradient camera can be used to replace a traditional sensor, should account for the task at hand and its accuracy constraints. Note that we did not investigate architectures that may better fit this type of data, and which may have an impact on accuracy. We leave the investigation for future work, see also Section~\ref{sec:discussion}.

\subsection{Effects of Gradient Quantization}\label{sec:quantization}

In this paper, we study the tradeoff between power consumption and accuracy of binary gradient cameras. One factor that has a strong impact on both, is the number of bits we use to quantize the gradient, which, so far, we have assumed to be binary. Designing a sensor with a variable number of quantization bits, while allowing for low power consumption, could be challenging. However, graylevel information can be extracted from a binary gradient camera by accumulating multiple frames, captured at a high frame rate, and by combining them into a sum weighted by the time of activation~\cite{gallo2015gray}.

For the sensor proposed by Gottardi \etal~\cite{graincam}, the power of computing this multi-bit gradient can be estimated as:
\begin{equation}\label{eq:powerGray}
P = 2^{N}\cdot P_{\text{scan}} + P_{\text{deliver}},
\end{equation}
where $N$ is the number of quantization levels, $P_{\text{scan}}$ is the power required to scan all the rows of the sensor, and $P_{\text{deliver}}$ is the power to deliver the data out of the sensor, which depends on the number of active pixels~\cite{gallo2015gray}. Despite the fact that $P_{\text{deliver}}$ is an order of magnitude larger than $P_{\text{scan}}$, Equation~\ref{eq:powerGray} shows that the total power quickly grows with the number of bits.

\begin{figure}
\centering
\includegraphics[width=\columnwidth, trim={25mm 60mm 20mm 60mm}, clip]{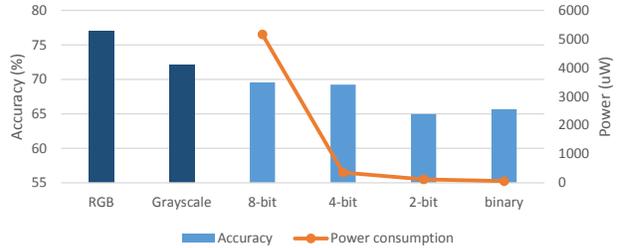}
\caption{Quantization vs power consumption vs accuracy tradeoff on CIFAR-10. Note the significant drop in power consumption between 8 and 4 bits, which is not reflected by a proportional loss of accuracy, see Section~\ref{sec:quantization}.}\label{fig:quantsweep}
\end{figure}

To study the compromise between power and number of bits, we simulated a multi-bit gradient sweep on CIFAR-10, and used Equation~\ref{eq:powerGray} to estimate the corresponding power consumption. Figure~\ref{fig:quantsweep} shows that going from a binary gradient to an  8-bit gradient allows for a 3.89\% increase in accuracy, but requires more than 80 times the power. However, a 4-bit gradient may offer a good compromise, seeing that it only requires 7\% of the power needed to estimate an 8-bit gradient (6 times the power required for the binary gradient), at a cost of only 0.34\% loss of accuracy. 
This experiment points to the fact that the trade-off between power consumption and accuracy can be tuned based on the requirements of the task, and possibly the use-case itself. Moreover, because in the modality described above $N$ can be changed at runtime, one can also devise a strategy where the quantization levels are kept low in some baseline operation mode, and increased when an event triggers the need for higher accuracy.

\section{Recovering Intensity Information from Spatial Binary Gradients}\label{sec:reconstruction}

In addition to the automated computer vision machinery, some applications may require a human observer to look at the data. An example is video surveillance: a low-power automatic system can run continuously to detect, for instance, a person coming in the field of view. When such an event is detected, it may be useful to have access to intensity data, which is more easily accessible by a human observer. One solution could be that a more power-hungry sensor, such as an intensity camera is activated when the binary gradient camera detects an interesting event~\cite{han2014glimpsedata}. Another solution could be to attempt to recover the grayscale information from the binary data itself. In this section, we show that this is indeed possible.

We outlined previous work on intensity reconstruction from temporal gradients in Section~\ref{sec:related}.
Currently available techniques, such as the method by Bardow \etal~\cite{bardow2016}, use advanced optimization algorithms and perform a type of Poisson surface integration~\cite{agrawal2006range} to recover the intensity information. However, they focus on the temporal version of the gradients. As a consequence, these methods can only reconstruct images captured by a moving camera, which severely limits their applicability to real-world scenarios.

To the best of our knowledge, there has been no work on reconstructing intensity images from a single binary spatial gradients image, in part because this problem does not have a unique solution. Capturing a dark ball against a bright background, for instance, would yield the same exact binary spatial gradient as a bright ball on a dark background. This ambiguity prevents the methods of surface integration from working, even with known or estimated boundary conditions. 

We take a deep learning approach to intensity reconstruction, so as to leverage the network's ability to learn priors about the data. For this purpose, we focus on the problem of intensity recovery from spatial gradients of faces. While we cannot hope to reconstruct the exact intensity variations of a face, we aim to reconstruct facial features from edge maps so that it can be visually interpreted by a human. Here we describe the network architecture we propose to use, and the synthetic data we used to train it. In Section~\ref{sec:prototype} we show reconstructions from real data we captured using a binary gradient camera prototype.

\begin{figure}
\centering
\includegraphics[width=\columnwidth]{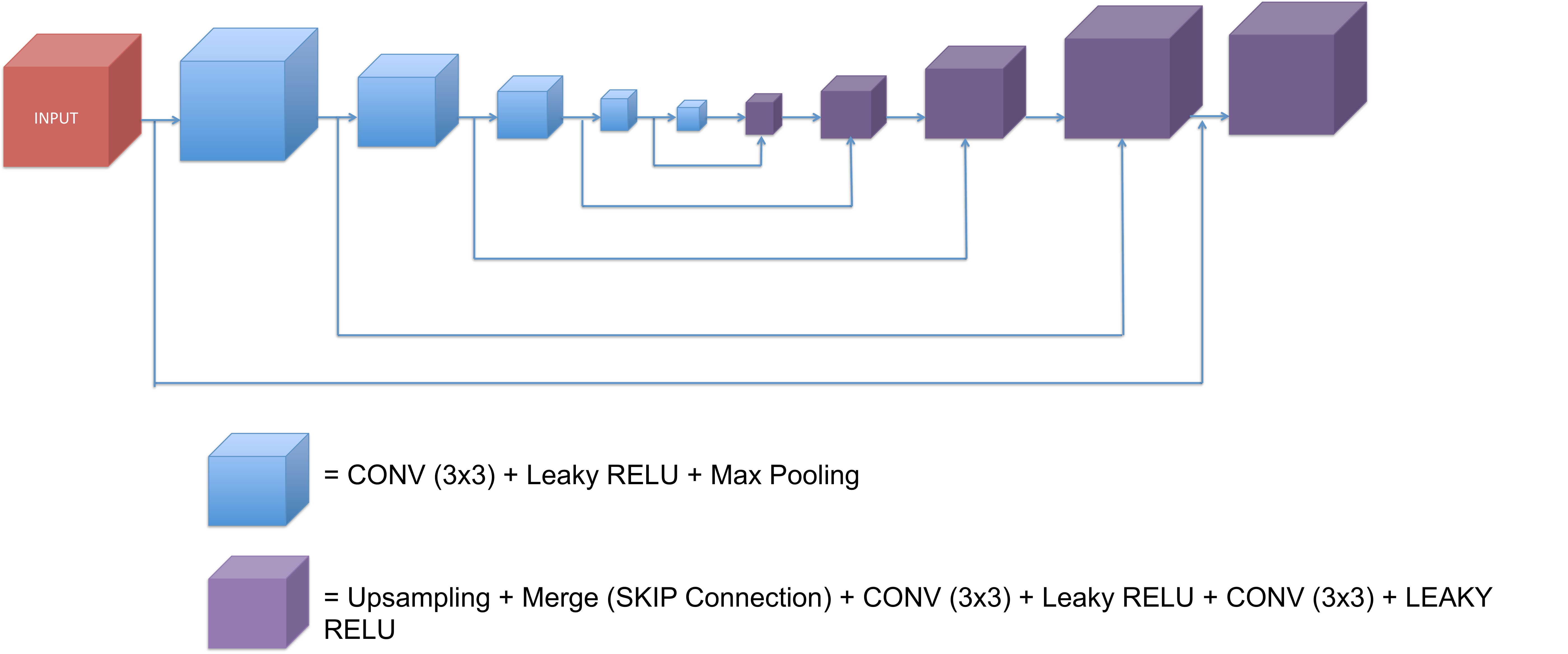}
\caption{The architecture of the autoencoder used to reconstruct intensity information from spatial binary gradient images.}\label{fig:autoencoder}
\end{figure}

Our network is inspired by the autoencoder architecture recently proposed by Mao \etal~\cite{mao2016convAE}. The encoding part consists of 5 units, each consisting of two convolutional layers with leaky ReLU nonlinearities followed by a max pooling layer. The decoding part is symmetric, with 5 units consisting of upsampling, a merging layer for skip connections that combines the activations after the convolutions from the corresponding encoder unit, and two convolutional layers. See Figure~\ref{fig:autoencoder} for our network structure. We trained this architecture on the BIWI and WIDER datasets. 

\begin{figure}
\centering
\makebox[0mm][s]{
\hspace{-2mm}
\includegraphics[width=0.99\columnwidth]{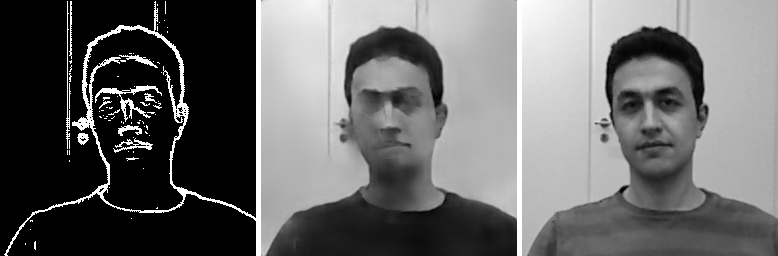}}
\hspace{-2.35mm}
\animategraphics[width=0.99\columnwidth]{15}{Figures/animations/BiwiFirst_}{0001}{0100}
 \caption{\textbf{Embedded animation} of the intensity reconstruction (middle pane) on the binary data (left pane) simulated from the BIWI dataset~\cite{biwi}. It can be viewed in Adobe Reader, or other media-enabled viewers, by clicking on the images. The ground truth is on the right.}\label{fig:intensityBIWI}
\end{figure}

For the BIWI dataset, we removed two subjects completely to be used for testing. Figure~\ref{fig:intensityBIWI} shows an embedded animation of the two testing subjects. As mentioned above, the solution is not unique given the binarized nature of the gradient image, and indeed the network fails to estimate the shade of the first subject's sweater. Nevertheless, the quality is sufficient to identify the person in the picture, which is surprising, given the sparseness of the input data.

\begin{figure*}
\centering
\includegraphics[width=\textwidth, trim={18mm 115mm 24mm 111mm}, clip]{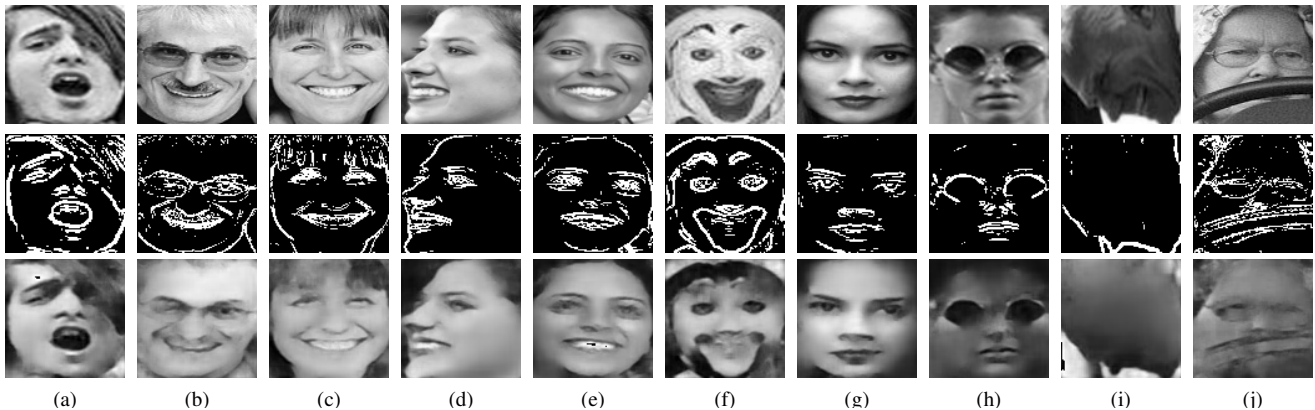}
\caption{Intensity reconstruction (bottom row) on the binary data (middle row) simulated from the WIDER dataset~\cite{wider}. The ground truth is in the top row. Note that our neural network is able to recover the fine details needed to identify the subjects. We observed that failure cases happen when the gradients are simply too poor (i) or the face is occluded (j).}\label{fig:intensityWIDER}
\end{figure*}

The WIDER dataset, does not contain repeated images of any one person, which guarantees that no test face is seen by the network during training. We extracted face crops by running the face detection algorithm described in Section~\ref{sec:benchmarks}, and resized them to 96x96, by either downsampling or upsampling, unless the original size was too small. Figure~\ref{fig:intensityWIDER} shows some results of the reconstruction. Note that the failure cases are those where the quality of the gradients is not sufficient (Figure~\ref{fig:intensityWIDER}(i)), or the face is occluded (Figure~\ref{fig:intensityWIDER}(j)). The rest of the faces are reconstructed unexpectedly well, given the input. Even for the face in Figure~\ref{fig:intensityWIDER}(j) the network is able to reconstruct the heavy makeup reasonably well.

\section{Experiments with a Prototype Spatial Binary Gradient Camera}\label{sec:prototype}

In this section we validate our findings by running experiments directly on real binary gradient images.
As a reminder, all the comparisons and tests we described so far were performed on data obtained by simulating the behavior of the binary gradient camera. Specifically, we based our simulator on Equation~\ref{eq:gs}, and tuned the threshold $T$ to roughly match the appearance of the simulated and real data, which we captured with the prototype camera described by Gottardi \etal~\cite{graincam}.
At capture time, we use the widest aperture setting possible to gain the most light, though at the cost of a shallower depth of field, which we did not find to affect the quality of the gradient image. We also captured a few grayscale images of the same scene with a second camera set up to roughly match the field of views of the two. Figure~\ref{fig:spatialgradcameras}, shows a comparison between a grayscale image and the (roughly) corresponding frame from the prototype camera. Barring resolution issues, at visual inspection we believe our simulations match the real data.

\subsection{Computer Vision Tasks on Real Data}

\begin{figure}
\centering
\subfloat[]{\includegraphics[width=.45\columnwidth]{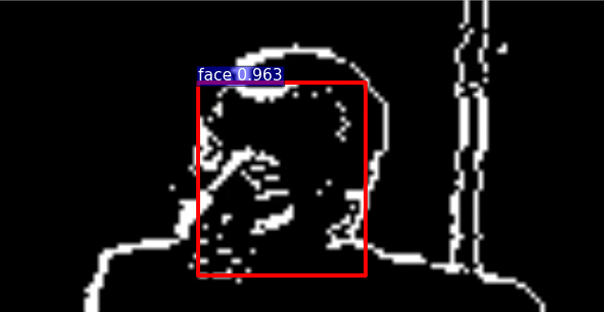}}\enskip
\subfloat[]{\includegraphics[width=.45\columnwidth,height=55px]{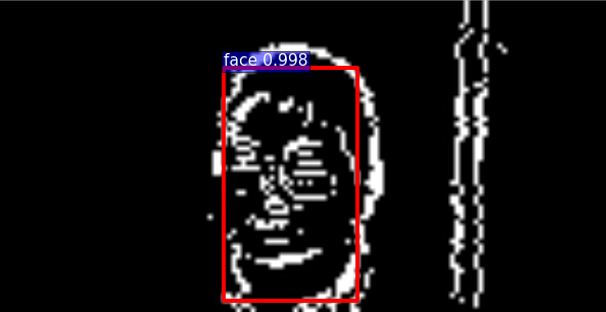}}\\
\vspace{-3.5mm}
\subfloat[]{\includegraphics[width=.45\columnwidth]{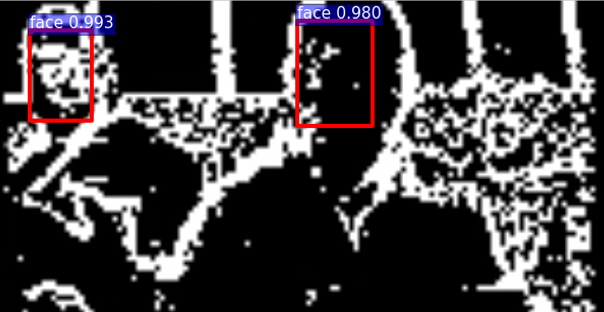}}\enskip
\subfloat[]{\includegraphics[width=.45\columnwidth]{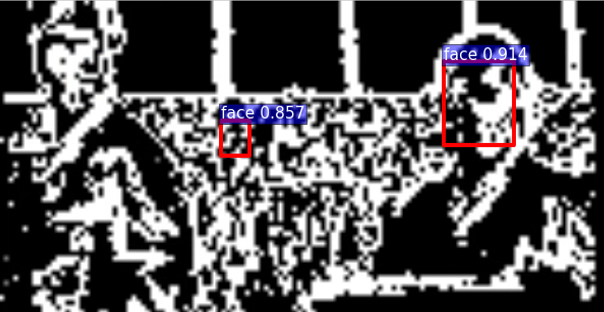}}
\caption{Face detection task on spatial gradient images captured with the camera prototype. The top and bottom rows show frames from an indoor and an outdoor sequence, respectively. The misdetection rate is significantly higher in outdoor sequences, as seen in insect (d).}\label{fig:prototypefacedetect}
\end{figure}

To qualitatively validate the results of our deep learning experiments, we ran face detection on binary gradient data captured in both outdoor and indoor environment. We could not train a network from scratch, due to the lack of a large dataset, which we could not capture with the current prototype---and the lack of ground truth data would have made it impossible to measure performance quantitatively anyway. We trained the network described in Section~\ref{sec:benchmarks} on simulated data resized to match the size of images produced by the camera prototype, and then we directly ran inference on the real data.We found that the same network worked well on the indoor scenes, missing a small fraction of the faces, and typically those whose pose deviated significantly from facing forward. On the other hand, the network struggled more when dealing with the cluttered background typical of the outdoor setting, where it missed a significant amount of faces. We ascribe this issue to the low spatial resolution offered by the prototype camera, which is only 128x64 pixels. However, this is not a fundamental limitation of the technology, and thus we expect it to be addressed in future versions.  Figure~\ref{fig:prototypefacedetect} shows a few detection results for both environment. 

\subsection{Intensity Reconstruction on Real Data}

\begin{figure}[h!]
\centering
\includegraphics[width=.24\columnwidth]{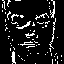}
\includegraphics[width=.24\columnwidth]{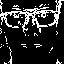}
\includegraphics[width=.24\columnwidth]{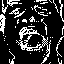}
\includegraphics[width=.24\columnwidth]{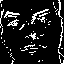}\\
\vspace{1mm}
\includegraphics[width=.24\columnwidth]{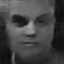}
\includegraphics[width=.24\columnwidth]{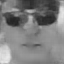}
\includegraphics[width=.24\columnwidth]{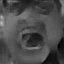}
\includegraphics[width=.24\columnwidth]{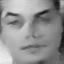}
\caption{Intensity reconstruction result inferred by the network described in Section~\ref{sec:reconstruction} and trained on the WIDER simulated data. The top row shows 64x64 face crops captured with the prototype camera, the bottom the corresponding reconstructed images. While the quality is not quite on par with the intensity reconstructions, it has to be noted that the resolution of the crops in Figure~\ref{fig:intensityWIDER}, is 96x96, \ie 1.5x larger.}\label{fig:intensityreal}
\end{figure}

Another qualitative validation we performed was intensity reconstruction from data captured directly with the camera prototype. We trained the network on synthetic data generated from the WIDER dataset, and performed forward inference on the real data. Once again, we could not perform fine-tuning due to the lack of ground truth data---the data from an intensity camera captured from a slightly different position, and with different lenses, did not generalize well. While the quality of the reconstruction is slightly degraded with respect to that of the synthetic data, the faces are reconstructed well. See Figure\ref{fig:intensityreal} for a few example. Note that despite the low resolution (these crops are 1.5 times smaller than those in Figure~\ref{fig:intensityWIDER}), the face features are still distinguishable.

Remember that here we are reconstructing intensity information from a single frame: we are not enforcing temporal consistency, nor we use information from multiple frames to better infer intensity. We find that the quality of the reconstruction of any single frame varies: some reconstructions from real data allow the viewer to determine the identity of the subject, others are more similar to average faces.

\section{Discussion}\label{sec:discussion}

To further decrease the power consumption in computer vision tasks, we could couple binary gradient images with binary neural networks. Recently, new architectures have been proposed that are use elementary layers (convolutions, fully connected layers) using binary weights, yielding an additional 40\% in power savings in computation~\cite{courbariauxbinarized}. We evaluated these binary neural networks (BNNs) on, MNIST, CIFAR-10, and SVHN~\cite{SVHN}. (The latter is a dataset of $\sim$100K house street numbers.) On MNIST, a 1.57\% error on gradient images increased to 2.23 \% error by employing a BNN. For CIFAR-10, a 11\% error on gradient images increased to 30\% with the BNN. Finally for SVHN, a 3\% error on binary gradient images increased to 12\% with the BNN. Thus, while there are considerable power savings from using a BNN, it is still an open question of how to couple these networks with binary gradient data from novel sensors. We leave this as an avenue for future work on end-to-end binary vision systems. 

We have conducted a thorough exploration of different computer vision tasks that can leverage binary gradient images. Certain tasks, such as object recognition and face detection, suffer more degradation in accuracy. Other tasks, such a gesture recognition, see an increase in accuracy. All with a significant power saving. In addition, we propose to use an autoencoder network to learn the prior distribution of a specific class of images to solve the under-constrained problem of recovering intensity information from binary spatial edges.

\section*{Acknowledgements}
We would like to thank Pavlo Molchanov for training and testing our data with the model described in~\cite{molchanov2016nvgesture}, and Massimo Gottardi for lending us the camera prototype used in this submission paper.

{\small
\bibliographystyle{ieee}
\bibliography{egbib}
}

\end{document}